\pdfoutput=1

\documentclass[11pt]{article}

\usepackage[]{acl}

\usepackage{times}
\usepackage{latexsym}

\usepackage[T1]{fontenc}

\usepackage[utf8]{inputenc}

\usepackage{microtype}

\usepackage{inconsolata}

\usepackage{graphicx}
\usepackage{multirow}
\usepackage{multicol}
\usepackage{subcaption}
\usepackage{enumitem,kantlipsum}
\usepackage[normalem]{ulem}
\usepackage{tabularx}
\usepackage{color, colortbl}
\usepackage{colortbl}
\usepackage{xcolor}
\usepackage{todonotes} 
\usepackage{rotating}
\usepackage{pbox}
\usepackage{booktabs} 
\usepackage{amssymb}
\usepackage{pifont}

\usepackage{tabu}
\usepackage{fontawesome}
\newcommand{\sq}{\faCheckSquare}

\definecolor{shadowcolor}{rgb}{0.8, 0.8, 0.8}

\newcommand{\areval}{ArAIEval}
\newcommand{\arpro}{\textbf{\textit{ArPro}}}

\title{ArAIEval Shared Task: Propagandistic Techniques Detection in \\Unimodal and Multimodal Arabic Content}
  \author{Maram Hasanain$^1$$^*$, Md. Arid Hasan$^2$\thanks{~~Equal contribution.}, Fatema Ahmed$^1$, Reem Suwaileh$^3$,\\
  \textbf{Md. Rafiul Biswas$^3$, Wajdi Zaghouani$^4$, Firoj Alam$^1$}\\
  $^1$Qatar Computing Research Institute, HBKU, Doha, Qatar \\
  $^2$ University of New Brunswick, Fredericton, NB, Canada \\  
  $^3$Hamad Bin Khalifa University, HBKU, Doha, Qatar \\  
  $^4$Northwestern University in Qatar, Education City, Doha,
Qatar \\  
  \texttt{\{mhasanain,rsuwaileh,fakter,mdbi30331,fialam\}@hbku.edu.qa}\\
\texttt{wajdi.zaghouani@northwestern.edu}, \texttt{arid.hasan@unb.ca}\\
  \texttt{\url{https://araieval.gitlab.io/}}
  \\}

\begin{document}
\maketitle
\begin{abstract}
We present an overview of the second edition of the \areval{} shared task, organized as part of the ArabicNLP 2024 conference co-located with ACL 2024. In this edition, \areval{} offers two tasks: ({\em i})~detection of propagandistic textual spans with persuasion techniques identification in tweets and news articles, and ({\em ii})~distinguishing between propagandistic and non-propagandistic memes. A total of 14 teams participated in the final evaluation phase, with 6 and 9 teams participating in Tasks 1 and 2, respectively. Finally, 11 teams submitted system description papers. Across both tasks, we observed that fine-tuning transformer models such as AraBERT was at the core of the majority of the participating systems. We provide a description of the task setup, including a description of the dataset construction and the evaluation setup. We further provide a brief overview of the participating systems. All datasets and evaluation scripts are released to the research community.\footnote{\url{https://araieval.gitlab.io/}} We hope this will enable further research on these important tasks in Arabic. 
\end{abstract}



\section{Introduction}
\label{sec:introduction}


Online media has become a primary channel for information dissemination and consumption, with numerous individuals considering it their main source of news~\cite{perrin2015social}. While online media (including news and social media platforms) offers a plethora of benefits, it is periodically exploited by malicious actors aiming to manipulate and mislead a broad audience. They often engage in sharing inappropriate content, misinformation, and disinformation \citep{alam-etal-2022-survey,ijcai2022p781}. Among the various forms of misleading and harmful content, propaganda is another communication tool designed to influence opinions and actions to achieve a specific objective. Propaganda is defined as a form of communication that is aimed at influencing the attitude of a community toward some cause or position by presenting only one side of an argument, which is achieved by means of well-defined rhetorical and psychological devices~\cite{InstituteforPropagandaAnalysis1938}. It is often biased or misleading in nature and is used to promote a particular political cause or point of view.  

News reporting in the mainstream media channels often use persuasion techniques to promote a particular editorial agenda. In different communication channels, propaganda is conveyed through the use of diverse persuasion techniques~\cite{Miller}, which range from leveraging the emotions of the audience, such as using \textit{emotional technique} or logical fallacies such as \textit{straw man} (misrepresenting someone's opinion), hidden \textit{ad-hominem fallacies}, and \textit{red herring} (presenting irrelevant data). 

In the past years, propaganda has been widely used on social media to influence and/or mislead the audience, which became a major concern for different stakeholders including social media platforms and government agencies. To combat the proliferation of propaganda on online platforms, there has been a significant surge in research in recent years. The aim is to automatically identify propagandistic content in textual, visual, and multimodal modalities, such as memes \cite{chen2023multimodal,SemEval2021-6-Dimitrov,EMNLP19DaSanMartino}. These efforts mainly focused on English, but with increased interest in other languages too.  

We enrich the Arabic AI research on that problem, by organizing a shared task on fine-grained propaganda techniques detection for Arabic, which attracted many participants~\cite{propaganda-detection:WANLP2022-overview}. We followed that by organizing the \textbf{Ar}abic \textbf{AI} \textbf{Eval}uation (\textbf{\areval}) shared task at ArabicNLP 2023, targeting binary and multilabel propaganda detection in Arabic text, with  14 and 8 teams participating in these subtasks, respectively~\cite{hasanain-etal-2023-araieval}.

Following the success of our previous shared tasks, and given the interest from the community in further pushing research in this domain, we organize the second edition of the \areval{}  shared task covering the following two tasks: {\em (i)} detection of propagandistic textual spans with persuasion techniques identification (unimodal), and {\em (ii)} distinguishing between propagandistic and non-propagandistic memes (multimodal). 

This edition of the shared task attracted good participation with a total of 45 registrations. Fourteen teams submitted their systems, and 11 teams submitted system description papers. For all tasks, majority of participating systems utilized transformer-based models, and several of them applied data augmentation techniques. All systems outperformed the respective random baselines. 

In the remainder of this paper, we define each of the two tasks, describe the manually constructed Arabic evaluation datasets, and provide an overview of the participating systems and their official scores.

\section{Related Work}
\label{sec:related_work}
\subsection{Persuasion Techniques}
The detection of persuasion techniques has brought significant attention across multiple research domains, including natural language processing (NLP), computational linguistics, psychology, and communication studies. Researchers have developed various approaches to automatically identify persuasive elements in digital content, leveraging advancements in NLP and machine learning. 

One prominent approach focuses on the linguistic analysis of persuasive texts. For instance, \citet{hidey2017analyzing} proposed a framework for identifying persuasion techniques in online discussions by analyzing rhetorical strategies such as appeals to emotion, authority, and logic. Their study demonstrated the potential of NLP techniques to classify and detect persuasive elements with high accuracy.

Building on this foundation, researchers have explored the use of machine learning models to enhance the detection of persuasion techniques. \citet{habernal2015exploiting} introduced the ArgumenText system, which applies supervised learning to classify argument components and identify persuasion techniques in user-generated content. This system showed promising results in distinguishing between different types of persuasive appeals.

In the domain of social media, the detection of persuasion techniques has been particularly challenging due to the diverse and dynamic nature of user interactions. One notable study by \citet{morstatter2018alt} utilized deep learning models to detect persuasion techniques in tweets related to political campaigns. By analyzing both textual features and user metadata, the researchers were able to identify persuasive content employed by political actors and their impact on public opinion.

The advent of a large language model has paved a new direction in persuasion techniques detection. \citet{hasanain2024can} developed \arpro{} containing 8K paragraphs labeled at the text span level for 23 propagandistic techniques and used it to evaluate the performance of GPT-4 in fine-grained propaganda detection. GPT-4 model has also shown improved performance when using a large-scale dataset with annotations from human annotators of varying expertise and providing more information to the model as prompts \cite{hasanain2023large}. 

\subsection{Multimodal Detection}
Multilingual and multimodal detection of persuasion techniques has also gained attention in recent years. 
\citet{park2016multimodal} created a multimodal corpus including acoustic, verbal, and visual features of 1,000 movie review videos to predict persuasiveness. \citet{fang2022emotion} embedded text and images using different networks and fused the multi-modal embeddings. A split-and-share module with multi-level representations then processes these fused features to improve the detection performance of persuasive techniques.

\citet{nojavanasghari2016deep} introduced a deep neural network architecture for predicting persuasiveness by combining three modalities: visual, acoustic, and text modalities. The multimodal fusion model outperforms unimodal models and can handle limited labeled data effectively. Another significant contribution to the field is the work by \citet{chen2017multimodal} on multimodal sentiment analysis, which is closely related to the detection of persuasive content.

\subsection{Related Shared Tasks}
Several shared tasks have been organized to engage the research community in persuasion techniques detection.  
The shared task organized by \citet{barronoverview} at the CLEF 2022 conference focused on detecting propaganda and persuasion techniques in news articles across multimodal data. Similarly, SemEval-2021 Task 6 \cite{SemEval2021-6-Dimitrov} focused on the detection of persuasion techniques in multimodal content, such as memes and videos, combining textual and visual analysis. 
The shared task in SemEval-2023 Task 3 \cite{piskorski-etal-2023-semeval} consists of 23 persuasion techniques detection  and has multiclass and multilabel classification setups focusing on the paragraph level.

Along with such initiatives, we have primarily focused on Arabic content. The propaganda detection shared task, co-located with WANLP 2022, was mainly focused on tweets in both binary and multilabel settings \cite{propaganda-detection:WANLP2022-overview}. Following the success of that task, we the ArAIEval shared task at Arabic NLP 2023, aimed to identify persuasive elements in Arabic tweets and news articles \cite{hasanain-etal-2023-araieval} in a multilabel setting. This year, we have expanded our dataset to include multimodal content in form of memes. We also focus on fine-grained propaganda detection at the span-level. This can motivate developing more robust models for detecting persuasion techniques in various Arabic media contexts.

\section{Task 1: Unimodal (Text) Propagandistic Technique Detection}
\subsection{Task Definition} 
The objective of this task is to develop and evaluate systems capable of detecting specific propagandistic techniques within textual content. The task is defined as follows. 
Given a multigenre text snippet (a news paragraph or a tweet), the task is to detect the propaganda techniques used in the text together with the exact span(s) in which each propaganda technique appears. This is a multilabel sequence tagging task.

\subsection{Data}
The dataset for this task covers two genres, tweets and paragraphs extracted from news articles. We collect raw data to annotate as follows.

\noindent\textbf{Tweets:}
We start from the same tweets dataset collected from Twitter accounts of Arabic news sources, as described in the previous editions of the shared task~\cite{propaganda-detection:WANLP2022-overview,hasanain-etal-2023-araieval}. In addition to that, we collect a fresh set of tweets targeting the current events in Gaza. The tweet set was collected by searching through the Twitter search API
using manually developed Arabic keywords.

\noindent\textbf{News paragraphs:}
The news paragraphs were sourced and annotated following the same approach followed to develop \arpro{} dataset~\cite{hasanain2024can}. News paragraphs are extracted from two collections of new  articles: \emph{(i)} AraFacts, and \emph{(ii)} a large-scale in-house collection. The \textbf{AraFacts} dataset~\cite{ali2021arafacts} contains true and false Arabic claims verified by fact-checking websites, and each claim is associated with online sources propagating or negating the claim. We only keep Web pages that are from news media in the set (e.g., www.alquds.co.uk). We automatically parsed the news articles and selected paragraphs for the annotation. As for the \textbf{in-house collection}, it consists of $600K$ news articles from over 400 news media outlets. The dataset is very diverse, covering 14 different broad topics and a wide coverage of Arabic news media.

\begin{figure}
    \centering
    \includegraphics[width=1\linewidth]{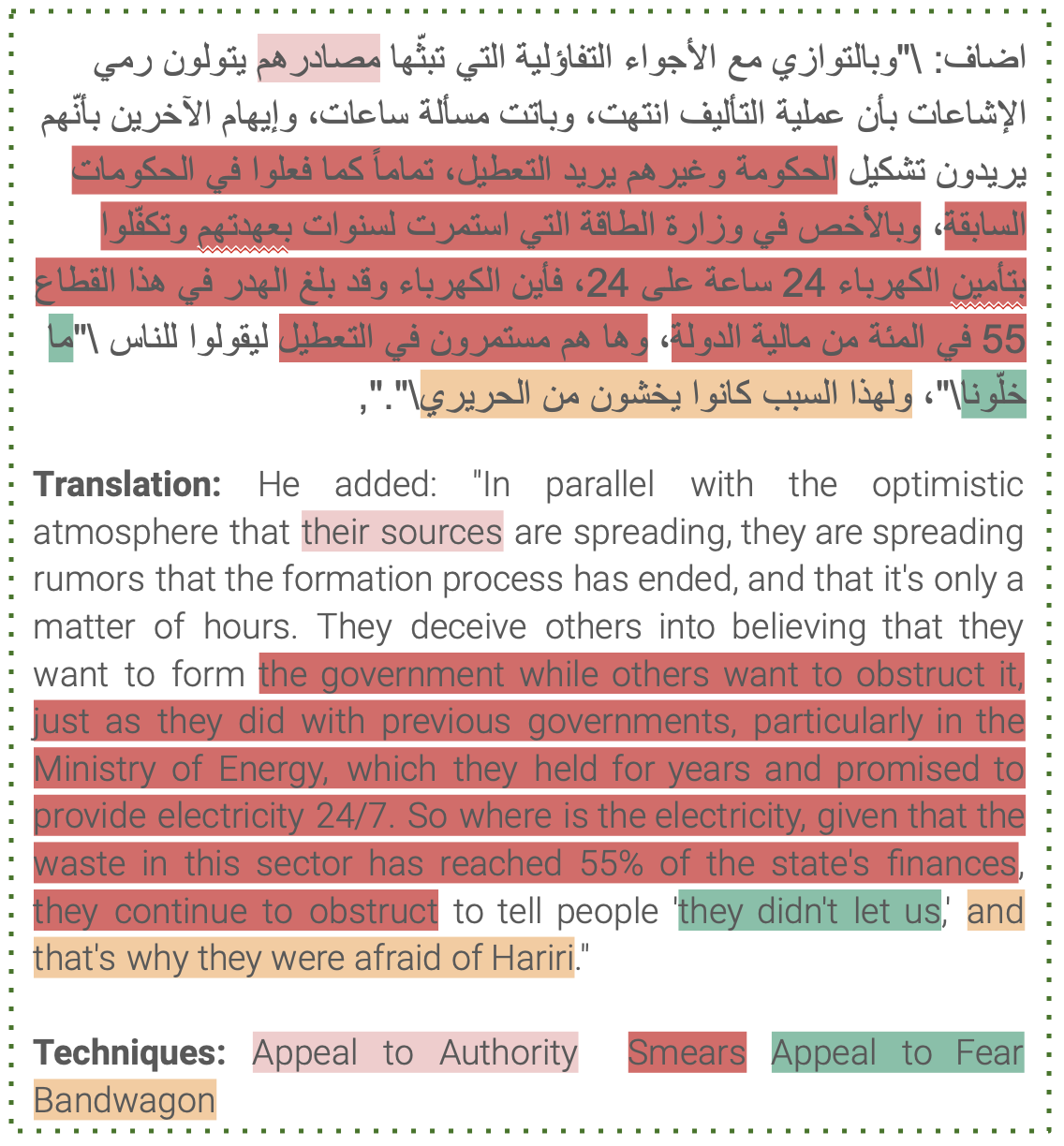}
    \caption{An example of a news paragraph annotated with propagandistic techniques.}
    \label{fig:prop_example}
\end{figure}

\paragraph{Data annotation:}

The annotation process includes two phases as detailed in previous studies~\cite{hasanain2024can,hasanain2023large}. The two annotation phases consist of: (\emph{i})~three annotators independently annotate the same text snippet by persuasion techniques on the span level, through an annotation interface designed for the task and (\emph{ii})~two consolidators (expert annotators) discuss each instance and finalize the gold annotations. Since annotations are at the fragment level, it might happen that an annotation is spotted by only one annotator. 
In order to train the annotators, we provide clear annotation instructions with examples and ask them to annotate a sample of text (tweets, or paragraphs from news articles). Then, we revise their annotations and provide feedback. In Figure \ref{fig:prop_example}, we provide an example of span annotation along with its English translation, which shows different text spans annotated with propagandistic techniques. 

The annotation taxonomy consists of a set of 23 persuasion techniques that is adopted from existing research~\cite{piskorski-etal-2023-semeval}. We should note here that multiple techniques can be found in the same text snippet even with an overlap. Below we give an example subset of the persuasion techniques, and briefly summarize them:
\begin{enumerate}
[leftmargin=*,noitemsep,topsep=0pt,leftmargin=*,labelwidth=!,labelsep=.5em]
\item \textbf{Name calling or labeling:} labeling the object of the propaganda campaign as something that the target audience fears, hates, finds undesirable, or loves, praises.
\item \textbf{Appeal to Fear, Prejudice:} building support or rejection for an idea by instilling fear or repulsion towards it, or to an alternative idea.
\item \textbf{Strawman:} giving the impression that an argument is refuted, whereas the real subject of the argument was not addressed or refuted, but instead was replaced with a different one.
\end{enumerate}


\paragraph{Data splits:}
The full set of annotated paragraphs is divided into three subsets: train, development, and test, using a stratified splitting approach to ensure that the distribution of persuasion techniques is consistent across the splits. For the tweets set, we split the full annotated tweet set from the previous edition of the lab~\cite{propaganda-detection:WANLP2022-overview} into train and development subsets, while the test set is annotated for this shared task. Finally, we construct the multi-genre subsets for the task by merging the sets of paragraphs and tweets. The full dataset is $\sim$9K snippets, randomly split into $\sim$78\%, $\sim$11\% and $\sim$11\% training, development and test splits, respectively.

\paragraph{Statistics:}
In Tables \ref{tab:dataset_task1} we show the distribution of labels across splits for Task 1. It reports that the techniques \textit{loaded language}, \textit{name calling}, and \textit{exaggeration-minimisation} are the most frequent among all techniques. Overall, the label space is highly imbalanced, which is typical for these types of tasks, as seen in previous studies~\cite{EMNLP19DaSanMartino,SemEval2021-6-Dimitrov}. 

\begin{table}[]
\centering
\setlength{\tabcolsep}{2pt} 
\scalebox{0.85}{%
\begin{tabular}{@{}lrrr@{}}
\toprule
\textbf{Technique} & \multicolumn{1}{l}{\textbf{Train}} & \multicolumn{1}{l}{\textbf{Dev}} & \multicolumn{1}{l}{\textbf{Test}} \\ \midrule
Appeal to Authority & 218 & 28 & 13 \\
Appeal to Fear-Prejudice & 180 & 32 & 7 \\
Appeal to Hypocrisy & 108 & 16 & 7 \\
Appeal to Popularity & 46 & 4 & 3 \\
Appeal to Time & 54 & 6 & 8 \\
Appeal to Values & 116 & 24 & 5 \\
Causal Oversimplification & 310 & 38 & 21 \\
Consequential Oversimplification & 83 & 11 & 6 \\
Conversation Killer & 69 & 10 & 6 \\
Doubt & 293 & 44 & 26 \\
Exaggeration-Minimisation & 1077 & 141 & 133 \\
False Dilemma-No Choice & 74 & 9 & 5 \\
Flag Waving & 218 & 34 & 17 \\
Guilt by Association & 23 & 2 & 4 \\
Loaded Language & 8779 & 1073 & 1429 \\
Name Calling-Labeling & 2243 & 330 & 432 \\
Obfuscation-Vagueness-Confusion & 578 & 67 & 55 \\
Questioning the Reputation & 872 & 128 & 106 \\
Red Herring & 41 & 5 & 3 \\
Repetition & 144 & 17 & 17 \\
Slogans & 190 & 37 & 7 \\
Straw Man & 23 & 3 & 2 \\
Whataboutism & 26 & 5 & 1 \\ \bottomrule
\end{tabular}
}
\caption{Distribution of the persuasion techniques in Task 1 dataset across different splits.}
\label{tab:dataset_task1}
\end{table}

\begin{table*}[t]
\centering
\setlength{\tabcolsep}{2.3pt}
\scalebox{0.85}{%
\begin{tabular}{@{}l|lllll|l|l|lllll@{}}
\toprule
\multicolumn{1}{c}{\textbf{Team}} & \multicolumn{5}{|c}{\textbf{Transformer}} & \multicolumn{1}{|c}{\textbf{DL}} & \multicolumn{1}{|c}{\textbf{ML}} & \multicolumn{5}{|c}{\textbf{Misc.}} \\ \midrule
\multicolumn{1}{c}{\textbf{}} & \multicolumn{1}{|c}{\rotatebox{90}{\textbf{AraBERTv2}}} & \multicolumn{1}{c}{\rotatebox{90}{\textbf{MARBERT}}} & \multicolumn{1}{c}{\rotatebox{90}{\textbf{Arabic-BERT}}} & \multicolumn{1}{c}{\rotatebox{90}{\textbf{mBERT}}} & \multicolumn{1}{c}{\rotatebox{90}{\textbf{CAMeLBERT}}} & \multicolumn{1}{|c}{\rotatebox{90}{\textbf{BiLSTM/CNN/LSTM}}} & \multicolumn{1}{|c}{\rotatebox{90}{\textbf{RF/SVM/MNB}}} & \multicolumn{1}{|c}{\rotatebox{90}{\textbf{Data augm.}}} & \multicolumn{1}{c}{\rotatebox{90}{\textbf{Preprocessing}}} & \multicolumn{1}{c}{\rotatebox{90}{\textbf{Hyperparameter tuning}}} & \multicolumn{1}{c}{\rotatebox{90}{\textbf{Feature Engineering}}} & \multicolumn{1}{c}{\rotatebox{90}{\textbf{Model Layer}}} \\ \midrule
CUET\_sstm~\cite{araieval-arabicnlp:2024:task:CUET_sstm} & \sq & \sq & \sq &  &  & \sq & \sq & \sq & \sq &  &  &  \\
Mela~\cite{araieval-arabicnlp:2024:task1:mela} & \sq &  &  & \sq &  &  &  &  &  &  &  & \sq \\
MemeMind~\cite{araieval-arabicnlp:2023:task1:MemeMind} & \sq &  &  & \sq & \sq &  &  &  & \sq & \sq &  &  \\
Nullpointer~\cite{araieval-arabicnlp:2023:task1:nullpointer} & \sq &  &  &  &  &  &  &  & \sq & \sq & \sq &  \\
SussexAI~\cite{araieval-arabicnlp:2023:task1:sussexAI} & \sq &  &  &  &  &  &  & \sq &  &  & \sq &  \\
SemanticCuetSync~\cite{araieval-arabicnlp:2024:task:SemanticCuetSync} & \sq &  & \sq &  & \sq & \sq & \sq &  &  &  &  &  \\ \bottomrule
\end{tabular}
}
\caption{\textbf{Task~1} Overview of the approaches. DL: Deep Learning, ML: Classic Machine Learning.}
\label{tab:overview_approaches_task1}
\end{table*}

\begin{table}[]
\centering
\begin{tabular}{@{}lrr@{}}
\toprule
\multicolumn{1}{c}{\textbf{Team}}& \multicolumn{1}{c}{\textbf{Rank}} & \multicolumn{1}{c}{\textbf{Micro F1}} \\ \midrule
CUET\_sstm &1& 0.2995 \\
Mela & 2 & 0.2833 \\
MemeMind & 3 & 0.2774 \\
Nullpointer & 4 & 0.2541 \\
SussexAI & 5 & 0.1228 \\
SemanticCuetSync & 6 & 0.0783 \\
\rowcolor{shadowcolor} Baseline & & 0.0151 \\ \bottomrule
\end{tabular}
\caption{Official results for \textbf{Task 1}. Runs ranked by the official measure: modified micro F1.}
\label{tab:results_task1}
\end{table}

\subsection{Evaluation Setup}
\label{ssec:evaluation_setup}
The task was organized into two phases:
\begin{itemize}[noitemsep,topsep=0pt,leftmargin=*,labelwidth=!,labelsep=.5em]
    \item \textbf{Development phase}: we released the train and development subsets, and participants submitted runs on the development set through a competition on Codalab.\footnote{\href{https://codalab.lisn.upsaclay.fr/competitions/18111}
    {https://codalab.lisn.upsaclay.fr/competitions/18111}}    
    \item \textbf{Test phase}: we released the official test subset, and the participants were given a few days to submit their final predictions through the same Codalab competition. Only the latest submission from each team was considered official and was used for the final team ranking.
\end{itemize}

\noindent\textbf{Measures:}
Task 1 is a multilabel sequence tagging task. We measure the performance of the participating systems using a modified F$_1$ measure that accounts for partial matching between the spans across the gold labels and the predictions~\cite{propaganda-detection:WANLP2022-overview}.

\subsection{Results and Overview of the Systems}
\label{ssec:results}

A total of 6 teams submitted runs for the evaluation phase of the task. In Table~\ref{tab:overview_approaches_task1}, we provide an overview of the participating systems for which a description paper was submitted.  
As shown in Table ~\ref{tab:overview_approaches_task1}, fine-tuning pre-trained Arabic models, specifically AraBERT~\cite{antoun2020arabert}, multilingual BERT (mBERT)~\cite{text_bert}, Arabic-BERT~\cite{safaya-etal-2020-kuisail}, and CAMeLBERT~\cite{inoue2021interplay} is the most common system architecture.

In Table \ref{tab:results_task1}, we report the results of the participants' systems including a random baseline. Results are competitive among the top four teams, performance differences are relatively lower among them. All systems surpass the random baseline.




\noindent Team \textbf{CUET\_sstm} \cite{araieval-arabicnlp:2024:task:CUET_sstm} used data augmentation with the synonym replacement approach from the nlpaug library. The data was preprocessed and encoded with the BIO technique to transform it into a token classification problem for span detection. 
They experimented with fine-tuning different models, including transformer-based models such as MAREFA-NER and AraBERT; however, Arabic-BERT outperformed the other models. 

\noindent Team \textbf{Mela} \cite{araieval-arabicnlp:2024:task1:mela} fine-tuned mBERT on the training set and experimented with different hidden layers of the model. Their experiments showed that the 10th hidden layer yielded the best F1 score on the development set. 

\noindent Team \textbf{SemanticCuetSync}~\cite{araieval-arabicnlp:2024:task:SemanticCuetSync} experimented with different types of models. Firstly, they trained typical machine learning models, including: Logistic Regression, Support Vector Machine, and Multinomial Naive Bayes. Secondly, they also trained deep learning-based models such as CNN, CNN+LSTM, and CNN+BiLSTM. Lastly, several transformer-based models were fine-tuned: AraBERTv2, CAMeLBERT, and Arabic-BERT. Among all the models, CAMeLBERT resulted in the highest micro-F1 score.

\noindent Team \textbf{SussexAI}~\cite{araieval-arabicnlp:2023:task1:sussexAI} performed data augmentation on least represented labels to reduce label imbalance. They also generated synthetic data using random masking. They truncated `non-propaganda' tokens outside `propaganda' spans and reduced `non-propaganda' tokens. Finally, they fine-tuned the AraBERT transformer model with three dataset setups: augmented, non-augmented, and random truncation. 

\noindent Team \textbf{MemeMind}~\cite{araieval-arabicnlp:2023:task1:MemeMind} encoded each label and created BIO tags, modeling the problem as a token classification task. Then, they fine-tuned AraBERTv2 in two steps: trained only the classification layer for a few epochs and then fine-tuned the full model by updating all parameters. They tried different transformer models. Among them, AraBERTv2 showed the best performance.

\noindent Team \textbf{Nullpointer}~\cite{araieval-arabicnlp:2023:task1:nullpointer} applied different preprocessing techniques, such as filtering out Unicode characters. 
Then, they fine-tuned AraBERTv2 following two setups to model the task: \textit{(i)} model the problem as a token classification task, with different label aggregation applied overt tokens at prediction time to find per-word label, and \textit{(ii)} model the problem as a word classification task, representing each word using max-pooling of embeddings of its
constituent tokens.



\section{Task 2: Multimodal Propagandistic Memes Classification}
\label{sec:task2}

\subsection{Task Definition}
The goal of this task is to develop and evaluate systems capable of classifying multimodal propagandistic memes. Memes typically consist of a background image, and a layer of text that adds context, humour, or commentary to the image. The combination of the image and the text creates a specific message, joke, or commentary that is meant to be easily understood, relatable, and shareable.

The aim of this task is to foster research in the intersection of multimodal learning and propaganda detection, encouraging innovative solutions that effectively combine visual and textual information to identify and understand propagandistic contents. 
To account for the different modalities covered by this task, we model the task into three binary classification subtasks. 

\begin{itemize}[itemsep=0pt, topsep=0pt, partopsep=0pt, parsep=0pt]
    \item \textbf{Subtask 2A:} Given a text extracted from a meme, categorize whether it is propagandistic or not. 
    
    \item \textbf{Subtask 2B:} Given a meme (text overlayed image), detect whether the content is propagandistic.  
    
    \item \textbf{Subtask 2C:} Given multimodal content (text extracted from meme and the meme itself), detect whether the content is propagandistic.  
\end{itemize}

\begin{figure}[t]
    \centering
    \includegraphics[scale=0.4]{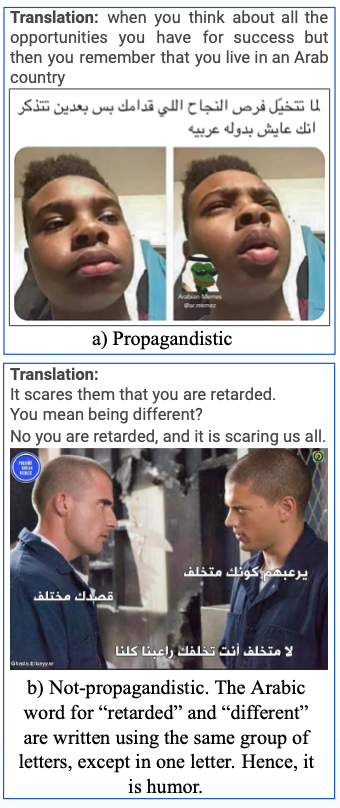}
    \caption{An example of memes with propagandistic and not-propagandistic categories.   
    }
    \label{fig:annotation_example_task}
\end{figure}

\begin{table}[]
\centering
\setlength{\tabcolsep}{2pt} 
\scalebox{0.95}{%
\begin{tabular}{@{}lrrrr@{}}
\toprule
\multicolumn{1}{c}{\textbf{Class labels}} & \multicolumn{1}{c}{\textbf{Train}} & \multicolumn{1}{c}{\textbf{Dev}} & \multicolumn{1}{c}{\textbf{Test}} & \multicolumn{1}{c}{\textbf{Total}} \\ \midrule
Not propaganda & 1,540 & 224 & 436 & 2,200 \\
Propaganda & 603 & 88 & 171 & 862 \\ \midrule
\textbf{Total} & 2,143 & 312 & 607 & 3,062 \\ \bottomrule
\end{tabular}
}
\caption{Distribution of Task 2 dataset.}
\label{tab:dataset_task2}
\end{table}

\begin{table*}[]
\centering
\setlength{\tabcolsep}{2.3pt}
\scalebox{0.8}{%

\begin{tabular}{@{}l|lll|lllllll|llllllllll|ll@{}}
\toprule
\multicolumn{1}{c}{\textbf{Team}}  & \multicolumn{3}{|c}{\textbf{Subtask}} & \multicolumn{7}{|c}{\textbf{Text Models}} & \multicolumn{10}{|c}{\textbf{Image Models}} & \multicolumn{2}{|c}{\textbf{Misc.}} \\ \midrule
 &  \rotatebox{90}{2A} & \rotatebox{90}{2B} & \rotatebox{90}{2C} & \rotatebox{90}{AraBERTv2} & \rotatebox{90}{MARBERT} & \rotatebox{90}{ARABERT} & \rotatebox{90}{QARiB} & \rotatebox{90}{CAMeLBERT} & \rotatebox{90}{GigaBERT} & \rotatebox{90}{BLOOMZ} & \rotatebox{90}{CLIP} & \rotatebox{90}{EfficientFormer} & \rotatebox{90}{ConvNeXt-tiny} & \rotatebox{90}{CAFormer} & \rotatebox{90}{EVA} & \rotatebox{90}{MaxViT} & \rotatebox{90}{SwinV2} & \rotatebox{90}{ResNet} & \rotatebox{90}{DenseNet} & \rotatebox{90}{LSTM} & \rotatebox{90}{Data aug.} & \rotatebox{90}{Preprocessing} \\ \midrule
AlexUNLP-MZ \cite{araieval-arabicnlp:2023:task2:alexUNLP} & \sq & \sq & \sq & \sq &  &  &  & \sq &  & \sq &  &  &  &  &  &  &  & \sq & \sq &  &  & \sq \\
ASOS \cite{araieval-arabicnlp:2024:task:ASOS} & &  & \sq &  & \sq & \sq & \sq &  &  &  &  &  &  &  &  & \sq & \sq & \sq &  &  &  & \sq \\
CLTL \cite{araieval-arabicnlp:2023:task1:CLTL} & \sq & \sq & \sq &  & \sq &  &  & \sq & \sq &  &  & \sq & \sq &  &  &  &  &  &  &  &  &  \\
MemeMind \cite{araieval-arabicnlp:2023:task2:MemeMind} & \sq & \sq & \sq & \sq & \sq &  & \sq & \sq &  &  &  &  &  & \sq & \sq &  &  & \sq &  &  & \sq &  \\
MODOS \cite{araieval-arabicnlp:2024:task:MODOS} &  &  & \sq &  &  &  &  &  &  &  & \sq &  &  &  &  &  &  &  &  & \sq &  &  \\ \bottomrule
\end{tabular}%
}
\caption{\textbf{Task~2} Overview of the approaches. DL: Deep Learning, Data aug: Data augmentation}
\label{tab:overview_approaches_task2}
\end{table*}

\begin{table}[tbh!]
\centering
\setlength{\tabcolsep}{2pt} 
\scalebox{0.95}{%
\begin{tabular}{@{}lrr@{}}
\toprule
\multicolumn{1}{c}{\textbf{Team}} &\multicolumn{1}{c}{\textbf{Rank}} & \multicolumn{1}{c}{\textbf{Macro F1}} \\ \midrule
\multicolumn{3}{c}{\textbf{Subtask 2A}} \\ \midrule
AlexUNLP-MZ & 1 & 0.787 \\
CLTL & 2 & 0.779 \\
MemeMind & 3 & 0.746 \\
DLRG & 4 & 0.739 \\
One\_by\_zero & 5 & 0.674 \\
Z-Index & 6 & 0.633 \\
\rowcolor{shadowcolor}Baseline & & 0.453 \\ \midrule
\multicolumn{3}{c}{\textbf{Subtask 2B}} \\ \midrule
CLTL & 1 & 0.711 \\
MemeMind & 2 & 0.664 \\
AlexUNLP-MZ	 & 3 & 0.659 \\
\rowcolor{shadowcolor}Baseline & & 0.475 \\ \midrule
\multicolumn{3}{c}{\textbf{Subtask 2C}} \\ \midrule
AlexUNLP-MZ	 & 1 & 0.805 \\
ASOS & 2 & 0.798 \\
CLTL & 3 & 0.798 \\
MemeMind & 4 & 0.797 \\
Team Engima & 5 & 0.753 \\
MODOS & 6 & 0.729 \\
Z-Index & 7 & 0.712 \\
\rowcolor{shadowcolor} Baseline & & 0.493 \\ \bottomrule
\end{tabular}
}
\caption{Official results for \textbf{Subtask 2A, 2B and 2C}. Runs ranked by the official measure: Macro F1.}
\label{tab:subtask_2}
\end{table}

\subsection{Dataset}
\label{ssec:dataset_t2}
The dataset consists of $\sim$3K memes annotated as propagandistic vs not-propagandistic, which were collected from different social media (e.g., Facebook, Twitter, Instagram and Pinterest). Each of them were annotated by three annotators and majority decision is considered as the final label. Texts from the memes were extracted using an off-the-shelf OCR,\footnote{\url{https://github.com/JaidedAI/EasyOCR}} followed by manual editing of the propagandistic memes. An example of such memes is provided in Figure~\ref{fig:annotation_example_task}. 
The distribution of propagandistic and not-propagandistic labels are $40\%$ and $60\%$, respectively. More details of this dataset can be found in \cite{alam2024armeme}.

\paragraph{Data splits:}
The dataset is split into 70\%, 10\% and 20\% for training, development, and test, respectively. 

\paragraph{Statistics:}
Table \ref{tab:dataset_task2} reports the distribution of labels for Task 2. The proportion of propagandistic memes is low, with a total of 28\%, which makes the task more challenging.

\subsection{Evaluation Setup}
\label{ssec:evaluation_setup_t2}
Similar to Task 1, we also conducted this task in two phases (i.e., development and test) as discussed in Section~\ref{ssec:evaluation_setup}. Systems were evaluated using macro F1 as the official measure.

\subsection{Results Overview of the Systems}
\label{ssec:results_t2}

A total of 6, 3 and 7 teams submitted runs for Subtask 2A, 2B and 2C, respectively with nine unique teams. Table~\ref{tab:overview_approaches_task2}gives an overview of the participating systems for which a description paper was submitted. Three out of five teams participated in all subtasks. Among the teams, fine-tuning transformer models such as MARBERT~\cite{abdul-mageed-etal-2021-arbert} and CAMeLBERT~\cite{inoue2021interplay} is the most popular architecture. As for the vision models ResNet was the most popular choice. 
In Table~\ref{tab:subtask_2}, we report the results and rankings for \textit{all} systems across all subtasks. All systems outperformed the random baselines by a large margin. Among the three modalities, performances in the text and multimodal modalities (Subtasks 2A and 2C) were relatively higher compared to those in the image-only modality.

\noindent Team \textbf{AlexUNLP-MZ}~\cite{araieval-arabicnlp:2023:task2:alexUNLP} used a Large Language Model (LLM) to extract features from the dataset. They considered `weighted loss' and `contrastive loss' while training the model. For text classification, they used the BLOOMZ model. For image classification, they tried using CNN-based ResNet and DenseNet architectures. For the multimodal data, they used a fusion of two architectures, such as BLOOMZ-1b1 and ResNet101.

\noindent Team \textbf{ASOS}~\cite{araieval-arabicnlp:2024:task:ASOS} used the MARBERT model for text classification and ResNet50 for image classification, and finally fused these two models for multimodality. 

\noindent Team \textbf{CLTL}~\cite{araieval-arabicnlp:2023:task1:CLTL} applied MARBERT, CAMeLBERT and GigaBERT~\cite{lan2020gigabert} for text classification. CAMeLBERT showed superior performance compared to the other models. 
 For the image modality, they examined two models: EVA~\cite{EVA} and CAFormer~\cite{yu2023metaformer}. For multimodality, they merged the embeddings of text and image using the multilayer perceptron technique.  

\noindent Team \textbf{MODOS}~\cite{araieval-arabicnlp:2024:task:MODOS} preprocessed raw data before feeding it into the model. Firstly, they used the Segment Anything Model~\footnote{\url{https://segment-anything.com/}} for image segmentation of the meme images. Then, they employed the state-of-the-art image encoder CLIP~\footnote{\url{https://github.com/openai/CLIP}} to extract the image embeddings. Finally, they used LSTM for multimodal classification.

\noindent Team \textbf{MemeMind}~\cite{araieval-arabicnlp:2023:task2:MemeMind} used synthetic text generated by GPT-4~\cite{openai2023gpt4} for text data augmentation. For image modality, they augmented data using DALL-E-2 and fine-tuned ResNet50, EfficientFormer (v2), and ConvNeXt-tiny architectures. For the multimodality, they fused the ConvNeXt-tiny and BERT.



\section{Conclusion and Future Work}
\label{sec:conclusion}
We presented an overview of the \areval{} shared task, which consists of two tasks: \textit{(i)}~detection of propagandistic textual spans with persuasion techniques identification in tweets and news articles, and \textit{(ii)}~distinguishing between propagandistic and non-propagandistic memes. 
The task attracted the attention of many teams: a total of 45 teams registered to participate during the evaluation phase, with 6 and 9 teams eventually making an official submission on the test set for tasks 1 and 2, respectively. Finally, 11 teams submitted task description papers. Task 1 aimed to identify the propaganda techniques used in multi-genre text snippets, including tweets and news articles. The task was to detect textual spans with propagandistic techniques. On the other hand, Task 2 aimed to detect propaganda in memes (multimodal content) in binary classification settings for different modalities. For both tasks, majority of the systems fine-tuned pre-trained transformer models. 

Future editions of the ArAIEval shared task will explore more complex tasks, such as 
incorporating Arabic dialects. Increasing the dataset size and focusing on low-prevalence class labels will be crucial to improve model robustness. For task 2, future work could involve labeling the dataset with specific propagandistic techniques.   

\section*{Limitations}
The datasets for both tasks are skewed in label distribution, making system development more challenging. To address this problem, increasing the data size with a focus on low-prevalence class labels could be beneficial. 
As for Task 2, we observe that the systems achieved relatively better performance, likely because the task was relatively simple in nature -- a binary classification setting. 

\section*{Ethical Considerations}
The ArAIEval shared task involves the development of models to detect propagandistic content in Arabic text and memes. While this research aims to combat the spread of misleading and manipulative information, it is essential to consider the potential ethical implications.

The datasets used in the shared task may contain biases, as propaganda is often subjective and can be influenced by cultural, political, and social factors. It is crucial to acknowledge these biases and strive for diverse and representative datasets to ensure the models developed are as unbiased as possible.

The models developed in this shared task could potentially be misused by malicious actors to create more sophisticated propaganda or to target specific individuals or groups. Therefore, we ask developers to be aware of this issue while deploying models. 



\section*{Acknowledgments}
The work of M. Hasanain, F. Ahmed, R. Suwaileh, W. Zaghouani and F. Alam is partially supported by NPRP 14C-0916-210015 from the Qatar National Research Fund, which is a part of Qatar Research Development and Innovation Council (QRDI). The findings achieved herein are solely the responsibility of the authors.

\bibliography{bib/propaganda,bib/participants}
\bibstyle{acl_natbib}




\end{document}